# UA-Net: Uncertainty-Aware Network for TRISO Image Semantic Segmentation

Kyle Lucke[1,+], Zuzanna Krajewska-Travar[2,+], Shoukun Sun[1,+], Lu Cai[2,*], John D. Stempien[2], Min Xian[1,*]

[1] University of Idaho, 1776 Science Center Drive, Idaho Falls, Idaho 83402, U.S.A
[2] Idaho National Laboratory, 1955 N. Fremont Ave., Idaho Falls, Idaho 83415, U.S.A
[*] corresponding authors. lu.cai@inl.gov and mxian@uidaho.edu
[+] these authors contributed equally to this work

**ABSTRACT**

Tristructural isotropic (TRISO)-coated particle fuels experience irradiation-induced dimensional changes and chemical reactions during high-temperature neutron irradiation. Post-irradiation materialography is used to understand physical processes that may impact overall fuel performance (e.g., coating integrity and fission product retention). Conventionally, experts manually evaluate the appearance and features of thousands of cross sections of these sub-mm-sized samples—a task that is both tedious and subjective. In this work, we propose Uncertainty-Aware Network (UA-Net), a novel deep learning-based framework that automatically segments five characteristic regions of TRISO fuel cross-sectional micrographs and characterizes the uncertainty map of the dense predictions. The segmentation model incorporates a multi-stage pretraining strategy that learns general image representations using a pretrained backbone on ImageNet; the full model was then fine-tuned using TRISO micrographs from different irradiation experiments, followed by another fine-tuning stage using the AGR-5/6/7 TRISO particle cross sections. Finally, a meta-model for uncertainty prediction is proposed and integrated to estimate the uncertainty map of the segmentation results, which is vital for an uncertainty-aware identification of small defects from TRISO cross-section images. The segmentation model was evaluated using class-wise Intersection over Union (IoU), mean IoU (mIoU), class-wise Precision, and mean Precision (mP). The evaluation was performed using a test set of 102 images, achieving an mIoU and mP of 95.5% and 97.3%, respectively. The meta-model achieved a specificity and sensitivity of 91.8% and 93.5%, respectively, indicating strong performance in misclassification detection. The model was also applied to TRISO particles from new images (i.e., ones not involved in the training and validation processes) and then evaluated qualitatively, thus demonstrating its high level of performance in extracting layer regions accurately.

## Introduction

### TRISO Background

Tristructural isotropic (TRISO)-coated particle fuel is designed for use in high-temperature gas-cooled reactors with operating temperatures in excess of 1000°C. Typically, TRISO fuel particles (Figure 1[a]) are less than 1 mm in diameter and consist of a fuel kernel (usually uranium oxide or a mixture of uranium carbide in uranium oxide) coated with a low-density carbon buffer layer, an inner pyrolytic carbon (IPyC) layer, a silicon carbide (SiC) layer, and an outer pyrolytic carbon (OPyC) layer (Figure 1[b]). Thousands of TRISO particles are typically overcoated in a graphitic "matrix" material, then formed into larger fuel elements such as cylindrical "compacts" or spherical "pebbles" (Figure 1[c]). A traditional large TRISO-fueled reactor (of several hundred MWt or more) may have millions of compacts or hundreds of thousands of pebbles and billions or trillions of TRISO particles in its core[1,2]. The US Advanced Gas Reactor (AGR) Fuel Development and Qualification Program is performing research and development on TRISO-coated particle fuel to support deployment of high temperature gas-cooled reactors (HTGRs)[3]. Four fuel irradiation experiments have been completed in the Advanced Test Reactor (ATR) at Idaho National Laboratory (INL). AGR-2 was the second irradiation test, and AGR-5/6/7 was the final qualification test of AGR fuel made entirely at the engineering scale.

Detailed microstructural analysis of the kernels and TRISO coatings of TRISO particles—both as-fabricated (unirradiated) and post-irradiation—is crucial for understanding fuel performance and the possible mechanisms behind in-service degradation. However, with thousands of particles being present in each compact or pebble, performing this analysis manually is a cumbersome process susceptible to subjectivity in terms of identifying layer boundaries. Oak Ridge National Laboratory developed a way to automatically analyze optical images of as-fabricated TRISO particles based on layer boundary detection, thereby significantly improving qualification of key parameters (e.g., layer thickness) for quality control[4]. Recently, machine-learning techniques such as deep convolutional neural networks (DCNNs)[5] and the context-ensembled refinement network (CERNet)[6] have also been used to segment the layers of cross-section images of as-fabricated TRISO particles. However, segmenting irradiated TRISO particles poses additional challenges due to irradiation-induced changes that may make layer interfaces more diffuse, result in fractures in layers, or cause debonding between layers. For example, irradiation can induce densification in the low-density buffer layer, potentially resulting in a gap between the buffer and IPyC layers (Figure 1[b]). It may also blur the lines between the kernel and buffer layer as they react, making even manual segmentation difficult. To address these challenges, we recently developed RU-Net[7] for TRISO layer segmentation, and it successfully outperformed several widely recognized convolutional neural network (CNN) based segmentation models, including U-Net[8], Residual U-Net[9], and Attention U-Net[10]. Nevertheless, RU-Net faced challenges whenever cross-section preparation and/or image quality were inadequate. This paper focuses on further improving the segmentation of TRISO layers, particularly under these types of challenging conditions.

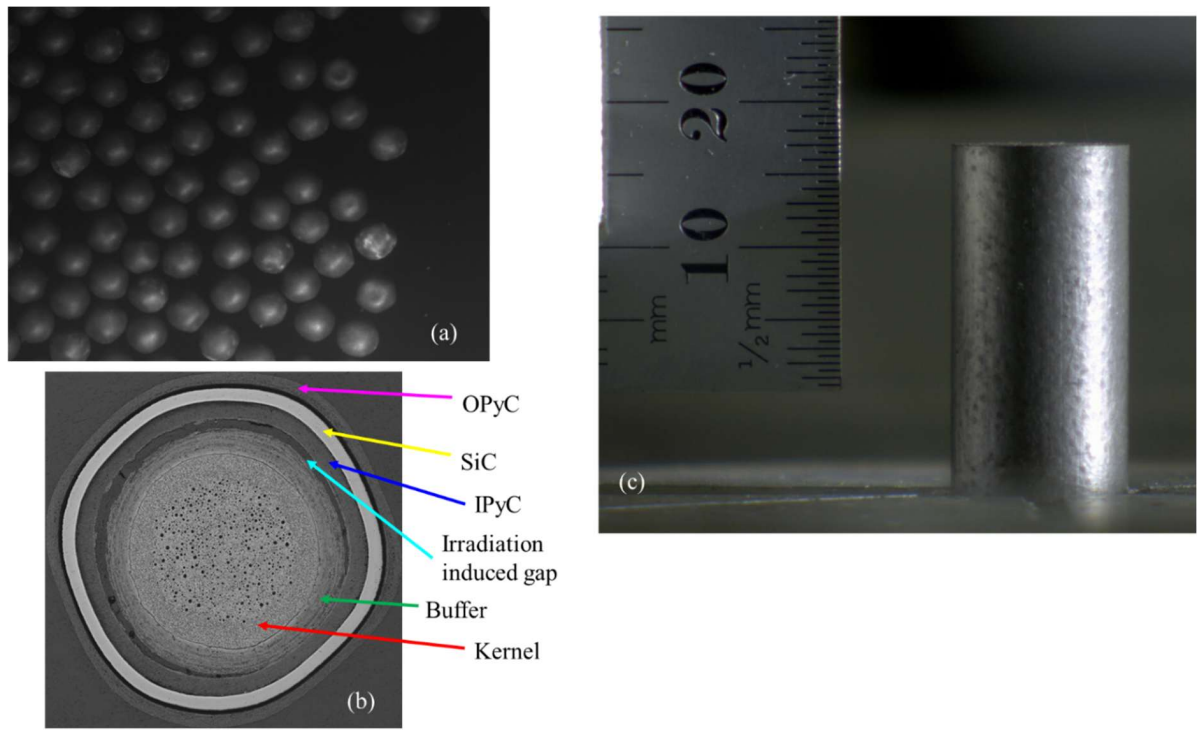

**Figure 1.** (a) Loose TRISO particles post-irradiation. (b) Optical microscopic image of a TRISO particle cross-section. (c) An AGR-5/6/7 fuel compact.

## Segmentation Approaches

Fully Convolutional Network[11] (FCN), was the first fully convolutional network for image segmentation. FCN provides the basis for all modern convolution-based segmentation networks. Its introduction of deconvolution layers (now colloquially known as transposed convolutions), which allow for learnable upsampling and skip connections that combine deeper and shallow features via addition, resulted in a more refined segmentation output. However, the model output is significantly smaller than the input, potentially leading to less precise boundary segmentation. U-Net[8] extends FCN and employs a symmetric encoder-decoder architecture with a larger number of feature maps and heavier use of skip connections. It consists of four encoder blocks, a bottleneck block, four decoder blocks, and a 1×1 convolution for obtaining the segmentation logits. Each encoder block consists of unpadded 3×3 convolutions followed by a 2×2 maximum pooling operation. The first convolution in each encoder block doubles the number of feature maps. The unpooled features from each encoder block are forwarded to the respective decoder block via skip connections. The bottleneck block consists of two unpadded 3×3 convolutions. Using transposed convolutions, the decoder block upsamples the feature maps from the previous decoder block by a factor of two, concatenates them with the encoder features, and applies two unpadded 3×3 convolutions. ReLU activations are applied after all convolution operations other than the final 1×1 convolution. The concatenation operation employed to combine the encoder and decoder features helps preserve spatial context and detail, resulting in better segmentation of fine details and boundaries. To further improve performance, several variants have been proposed[10,12–14]. However, some of them significantly increase the number of model parameters, increasing the risk of overfitting. One shortcoming of the original U-Net is the relatively simple encoder. While it could work well for some applications, a more advanced encoder could be necessary for other applications.

DeepLabV3[15] utilizes atrous convolutions to preserve feature map sizes in the model backbone as well as an atrous spatial pyramid pooling (ASPP) module. ASPP applies atrous convolutions with increasingly large dilation rates and a global average pooling layer, in parallel. The resulting feature maps are then concatenated, passed through two 1×1 convolutions, and bilinearly upsampled to produce the segmentation logits. The atrous convolutions in the model backbone help preserve feature map resolution, resulting in better segmentation of object boundaries and fine details. ASPP captures spatial information at different scales, resulting in better segmentation of various-sized objects, while the global average pooling layer encodes global context information to help resolve ambiguous image areas.

MobileNetV3[16], a lightweight classification model proposed for use with mobile phones, is an improvement over MobileNetV2[17] in terms of both accuracy and latency. It uses a hardware-aware neural architecture search for the model architecture and employs NetAdapt[18] to refine the individual layer definitions. The hardware-aware neural architecture search attempts to obtain a computationally optimal model architecture with respect to the targeted hardware (in their case, a cell phone). Their proposed segmentation head consists of an atrous convolution in the final encoder layer, a two-branch ASPP module, and a single skip connection from the MobileNetV3 encoder. In general, FCN, DeepLabV3, and MobileNetV3 produce segmentation maps that are smaller than the input size. This size difference is generally resolved by bilinearly upsampling the predicted segmentation maps, which could result in loss of fine details.

One shortcoming of CNN-based segmentation models is their inability to capture global context information. This could lead to difficulty in resolving ambiguous image areas. To mitigate this problem, transformer-based segmentation models have been proposed[19,20]. However, transformers have a weaker inductive bias than CNNs, and thus typically require large amounts of data to be effective. Hence, CNNs tend to perform better on small datasets[21]. This is important in the context of domain-specific applications where acquiring data is prohibitively expensive in terms of time or other resources.

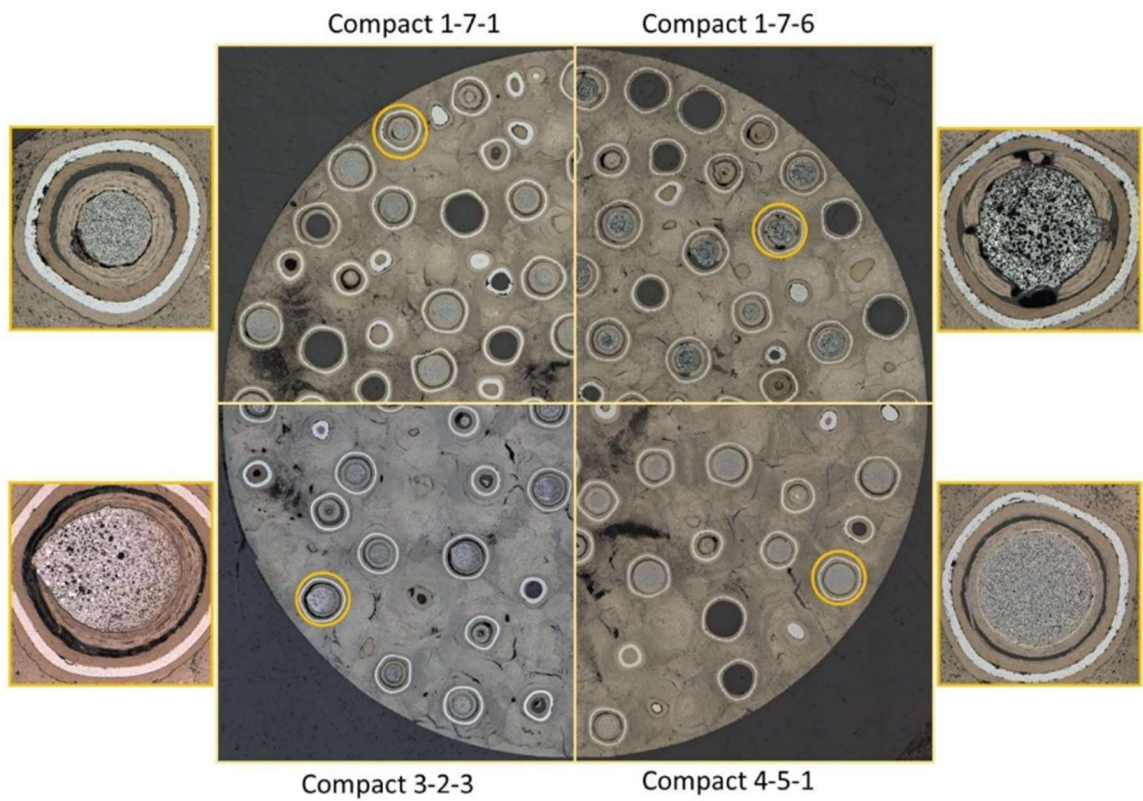

**Figure 2.** Disc mount representing one-quarter of each AGR-5/6/7 compact. Also shown is an example of TRISO particle taken from each compact.

To capture local and global context, hybrid CNN transformer segmentation models have also been proposed. These typically utilize either a transformer-based encoder with a CNN-based decoder[22–24], a dual-branch encoder (i.e., one CNN-based encoder and one transformer-based encoder)[25,26], or a mix of transformer and CNN layers[27]. However, large-scale pretraining is required for the transformer layers to be effective and may necessitate complex feature-fusion methods to obtain good performance.

**Uncertainty Quantification Approaches**

There has been a growing interest in recent years in quantifying the uncertainty associated with the predictions of deep learning models, typically known as uncertainty quantification (UQ). These measures of uncertainty can then be used to detect incorrect model predictions, or inputs that come from outside of the training distribution. In this work, we are concerned with detecting incorrect classifications, as failure to address them could lead to errors in downstream analysis or conclusions drawn from the segmentation results. While a large number of UQ methods exist, we only discuss some of the most prominent ones in the interest of brevity. Maximum softmax probability[28], which estimates uncertainty using the maximum predicted probability of the model, is a popular and simple to implement post-hoc baseline for UQ. However, it has been shown that deep-learning-based models can produce overconfident, erroneous predictions[29].

Deep ensembles[30] involve training several copies of the same predictive deep learning-based model. Once trained, uncertainty estimates can be obtained by taking the variance over the model predictions. They have been shown to outperform other UQ methods in several tasks[31]. However, they are prohibitively expensive for large-scale or real-time applications. This has prompted research into efficient ensemble methods[32–35] that are less computationally less expensive but tend to underperform a full deep ensemble.

Bayesian neural networks[36] (BNNs) model the network weights as a posterior distribution. Once trained, uncertainty estimates can be obtained by sampling from the posterior of the network weights and computing the variance of the model predictions. While grounded in well understood theory, BNNs can be difficult to train and are computationally expensive to obtain high quality uncertainty estimates from due to repeated sampling.

Meta-model-based UQ methods[37–40] present an effective post-hoc approach to UQ. They involve training an auxiliary model to predict the uncertainty associated with the task-model's predictions. The meta-model generally takes features generated by the task-model as input. These features can be the task model's predictive distribution, intermediate feature maps, or both. Due to their post-hoc nature, they can be used with any already trained predictive model without the need for model retraining. Moreover, because the auxiliary model tends to be significantly smaller than the predictive model, they are also computationally efficient, making them suitable for real-time or large-scale applications.

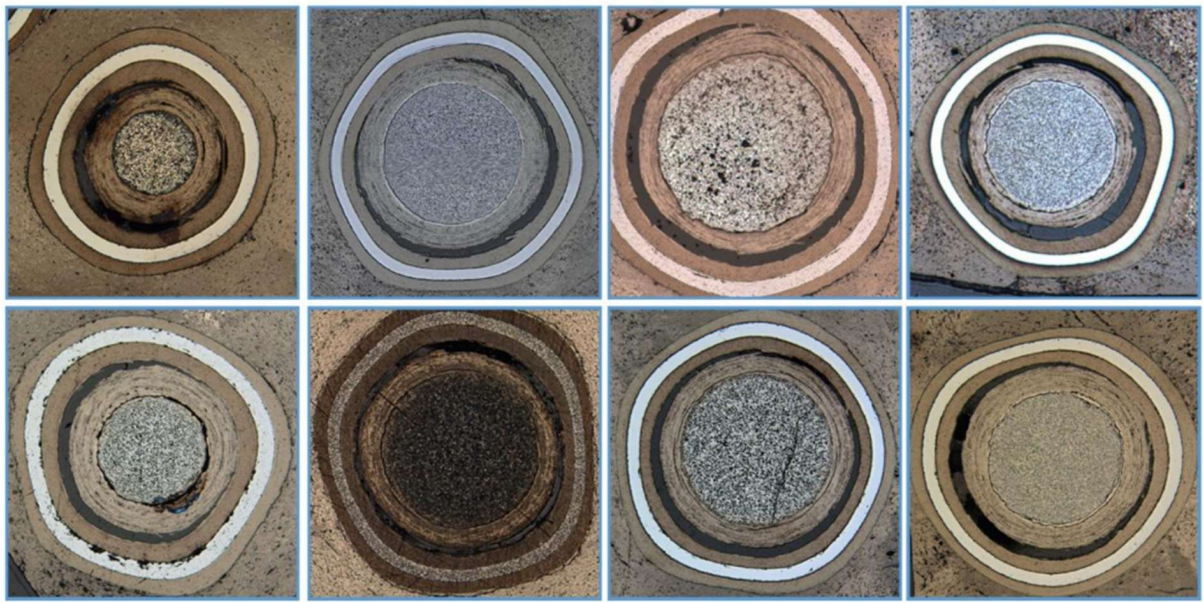

**Figure 3.** TRISO particle cross sections, representing varying quality of optical microscopy imaging.

## Data and Materials

### Data Collection and Preparation

Optical images were collected for two types of samples: AGR-2 TRISO particles and AGR-5/6/7 TRISO compacts. AGR-2 was the second irradiation experiment in the U.S. Advanced Gas Reactor (AGR) Fuel Development and Qualification program, and AGR-5/6/7 the final one[1]. Four irradiated AGR-2 compacts were electrolytically deconsolidated to free TRISO particles from the surrounding graphite matrix (see Figure 1(a)), whereas all the TRISO particles in the AGR-5/6/7 samples remained in the graphite matrix. The sample preparation method applied to the AGR-2 particles is detailed in [41,42], and involved embedding each particle in epoxy, followed by a series of iterative grinding and polishing steps, resembling the process described below for the AGR-5/6/7 compacts. A total of 2,171 optical images were collected for the AGR-2 particles, 1,764 of which were particles whose OPyC layer was removed. Note that removal of OPyC is standard practice during post-irradiation examination when measuring fission products but presents challenges for OPyC layer segmentation.

The TRISO particle fuel used in the AGR-5/6/7 program (i.e., UCO kernels, TRISO coatings, and fuel compacts) was produced at engineering scale at BWXT[43]. Thousands of TRISO particles were overcoated with graphitic matrix material and formed into cylindrical compacts. AGR-5/6/7 compacts are nominally 25.0 mm in length and 12.3 mm in diameter[44]. To identify irradiation-induced changes in the fuel kernel and TRISO coating layers, select compacts were subjected to destructive examination.

Selected compacts were sectioned both transversely and longitudinally via a low-speed diamond saw. From each compact, three mount cross sections (i.e., one disc and two sides) were obtained. These were then prepared via iterative grinding, with epoxy backpotting (for stabilizing the fragile structures in the cross section) and polishing prior to optical microscopy. Each mount was potted using Buehler Epoheat® epoxy aided with vacuum impregnation prior to heat curing. The mounts were then ground by first using 220-grit, then 500-grit, and finally 1200-grit Struers MD-Piano® grinding discs to remove the thick layer of epoxy covering the mounts. Post-grinding, all mounts were polished by starting with a 3-micron disc and moving up to 1-micron disc[45]. Figure 2 presents an example of a disc mount post-polishing, with TRISO particle cross sections having been exposed. Each quarter presented in that figure corresponds to the disc mount from a different irradiated AGR-5/6/7 compact. Micrographs of the TRISO particles circled in yellow are shown next to their respective compact mounts.

To verify the quality of the polished mounts, selected cross sections were investigated via optical microscopy. High-magnification, high-resolution color images of TRISO particles were obtained. Figure 3 presents examples of TRISO particle cross sections. The quality of the micrographs may differ due to variations in sample preparation or the conditions under which the images were collected. The microscope lighting was not perfectly uniform, and the samples not perfectly planar. Thus, montages may have a tiled appearance, and portions may be out of focus. The softer pyrolytic carbon (PyC) layers tend to polish faster than the harder SiC layer, creating relief when the SiC is at a higher plane than the PyC, such that it may cast a shadow onto the lower PyC layers. In some cases, artifacts from grinding show up on the samples as scratches, kernel pullout (e.g., row 2, column 2 of Figure 3), or pits in the SiC, hindering analysis of those images. Inconsistency in sample preparation or image acquisition not only challenged manual visual analysis but also made it harder to create a program for identifying the TRISO features of interest.

The micrographs were collected via 10×, 20×, and 50× objectives while traversing back and forth across the entire mount surface ("snake mode"). That mode resulted in obtaining hundreds of mount images representing parts of the TRISO particles. The micrographs were then merged via Adobe Photoshop to create a full montage of each mount cross section. Once the montages were created, each TRISO particle was assigned a serial number and examined in detail to identify certain morphologies and determine how frequently those morphologies occur. For machine-learning purposes, optical images with 10× and 20× magnification were used.

TRISO particles were analyzed for features such as cracks in the TRISO layers, kernel migration, kernel extrusion into buffer cracks, and debonding at the buffer-IPyC junction. The amount of material removed from each cross section during polishing was estimated to be approximately half a particle diameter (i.e., roughly 400–500 μm). In addition, some mounts were selected for repolishing, to improve the quality of SiC layer. The amount of material removed during repolishing is unknown. Each mount contains 50–100 TRISO particles per cross section. The number of particles per mount varied based on mount size and the packing fraction of the samples. AGR-5/6/7 compacts were fabricated with particle volume packing fractions of 40% and 25%. In this research, four AGR-5/6/7 compacts were investigated, giving us 985 TRISO particles to analyze.

**Layer Annotation**

For 510 of the particles, we used the Intelligent Image Annotation Platform (IIAP) to create detailed annotations of five image regions (i.e., kernel, buffer, IPyC, SiC, and OPyC), considering everything else (i.e., the graphite matrix and the gap between the buffer and IPyC layers) to be background. The IIAP provides an interactive tool for individually annotating each TRISO layer by placing a few seeds inside and outside the layer. Experts at Idaho National Laboratory verified the results and made necessary corrections. The annotated image regions serve as the ground truth for training and evaluating the proposed method.

**Uncertainty Ground Truth Generation**

To supervise the proposed uncertainty component of our method, we use a recently proposed soft-label scheme[39] as the ground truth uncertainty measure, which we refer to as SoftLabel. It can be calculated as

$$u_i = \begin{cases} P(y_i|x_i), & \text{if } \hat{y}_i = y_i \\ -(1 - P(y_i|x_i)), & \text{if } \hat{y}_i \neq y_i \end{cases}$$

where $u_i$ is the ground truth uncertainty value, $y_i$ is the ground truth class for the $i$th pixel, $\hat{y}_i$ is the corresponding segmentation model prediction, $P(y_i|x_i)$ is the probability predicted by the segmentation model that the $i$th pixel of the input $x$ belongs to the ground truth class $y$. By using separate equations for the correct and incorrect classifications, the meta-model can more accurately distinguish between them.

**Proposed Methods**

**Problem Formulation and Overall Framework**

UA-Net consists of two components: a segmentation model and a UQ model. The segmentation model produces segmentation maps for the TRISO particles, while the UQ model quantifies the uncertainty associated with the predictions of the segmentation model. The TRISO particle segmentation is formulated as a semantic image segmentation problem that partitions an image into non-overlapping regions by assigning a single semantic label to each pixel. Formally, the task is defined as a mapping from an input image $I$ to a segmentation map $Y$:

$$f: I(i) \to Y(i), i = 1, 2, \cdots, N \text{ and } Y(i) \in L = \{l_1, l_2, \cdots, l_n\} \tag{1}$$

where $I(i)$ is the value of the $i$th pixel, $N$ is the total number of pixels, $L$ is a set of class labels, and $n$ is the number of class labels. In deep learning, semantic image segmentation is equivalent to a pixel-wise mapping function $f$ that is defined by the architecture of a deep neural network. Finding $f$ is converted into a classification problem and is commonly formulated as a discriminative learning problem that maximizes the posterior probability ($p$) of the predicted segmentation mask ($y$), given the input image ($x$):

$$\max_{w_f} \log p_{w_f}(y|x) \tag{2}$$

where $w_f$ denotes the parameters of a discriminative model ($f$), and $x \in R^{H \times W \times 3}$ is the input image with height $H$ and width $W$. $y \in \{0, 1\}^{H \times W \times C}$ is a segmentation mask, where $C$ is the number of semantic categories.

The proposed segmentation model follows a U-Net[8] encoder-decoder structure with skip connections. The architecture is shown in Figure 4(a).

To quantify the uncertainty associated with the predictions of the segmentation model, we formulate the problem as pixel-level regression, where each pixel is assigned an uncertainty value. Formally, the task is defined as a mapping from the Softmax predictions of a segmentation model $P$ to an uncertainty map $U$:

$$g: P(i) \rightarrow U(i), i = 1, 2, \cdots, N \text{ and } U(i) \in [-1, 1] \tag{6}$$

where $P(i)$ is the probability distribution of the $i$th pixel predicted by the segmentation model. In deep learning, pixel-level regression is equivalent to a pixel-wise mapping function $g$ that is defined by the architecture of a deep neural network. Finding $g$ is commonly formulated as a discriminative learning problem that maximizes the posterior probability ($p$) of the predicted uncertainty map ($u$), given the predicted softmax probability distribution ($s$):

$$\max_{w_g} \log p_{w_g}(u|s) \tag{7}$$

where $w_g$ denotes the parameters of the parameters of a discriminative model ($g$), $s \in R^{H \times W \times C}$ is the probability distribution predicted by the segmentation model, and $u \in [-1, 1]^{H \times W}$ is the predicted uncertainty distribution.

In this work, we use a meta-model-based approach to quantify the uncertainty of predictions made by the segmentation model with the goal of identifying incorrectly segmented pixels. The proposed meta-model follows a standard U-Net[8] encoder-decoder structure with skip connections. The architecture is shown in Figure 4(b).

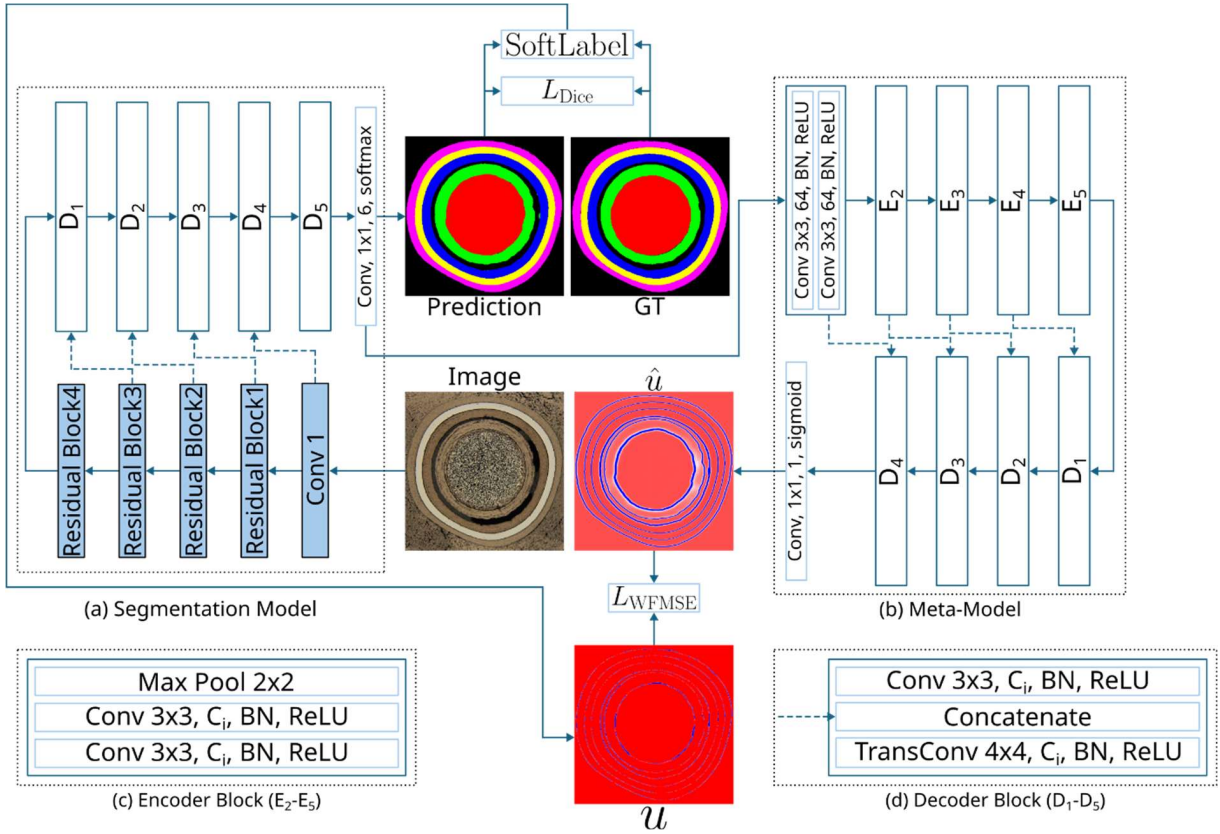

**Figure 4.** Architecture of the proposed framework. (a) Overall architecture of the segmentation model. (b) Overall architecture of the meta-model for UQ. (c) Encoder block structure. (d) Decoder block structure. Solid lines indicate the forward process of the model; dashed lines represent skip connections. In the segmentation map, red indicates the kernel layer, green the buffer layer, blue the IPyC layer, yellow the SiC layer, magenta the OPyC layer, and black the background layer. In the uncertainty map blue, white, and red indicate low, medium, and high uncertainty, respectively.

### Encoder and Decoder

The segmentation model encoder is a ResNet-152[46] backbone. The skip connection features are extracted from the first convolution as well as the outputs of the first, second, and third sets of ResNet layers.

The segmentation model decoder consists of five decoder blocks, followed by a convolution with a 1x1 kernel. Each decoder block consists of a transposed convolution, batch normalization (BN)[47], and ReLU[48] (TCBR) block, followed by a convolution, BN, and ReLU (CBR) block. The TCBR block takes the features from the previous decoder block as input and doubles the height and width of the features. It can be written as:

$$\text{TCBR}_i(D_{\text{out},i-1}) = \text{ReLU}\left(\text{BN}\left(TC_{4,4}(D_{\text{out},i-1})\right)\right), \quad (3)$$

where $D_{\text{out},i-1}$ is the output of the previous decoder block, $TC_{h,w}$ is a transposed convolution with a kernel of size $h, w$, and ReLU is the ReLU activation. The CBR block takes as input the output of the TCBR block and the features from the respective encoder block. It can be written as:

$$\text{CBR}_i(D_{\text{out},i-1}, S_i) = \text{ReLU}\left(\text{BN}\left(C_{3,3}([\text{TCBR}_i(D_{\text{out},i-1}), S_i])\right)\right), \quad (4)$$

where $S_i$ denotes the skip connections from the respective encoder layer, $[\cdot,\cdot]$ denotes channel-wise concatenation, and $C_{h,w}$ denotes a convolution with a kernel of size $h, w$. Note that the final decoder block does not use a skip connection. The decoder block architecture is shown in Figure 4(d). The number of channels (denoted $C_i$ in Figure 4(d)) for the decoder blocks $D_1, D_2, D_3, D_4$, and $D_5$ are 256, 128, 64, 32, and 16, respectively.

The meta-model encoder is standard U-Net encoder and consists of five encoder blocks. Each encoder block consists of a max pooling layer with a 2x2 kernel and two convolutions with a 3x3 kernel, where each convolution is followed by a BN layer and ReLU activation. The first encoder block does not have a max-pooling layer. The skip connection features are extracted from the outputs of each encoder block. The encoder block architecture is shown in Figure 4(c). The number of channels (denoted $C_i$ in figure 4(c)) for the encoder blocks $E_2, E_3, E_4, E_5$ are 128, 256, 512, and 1024, respectively. The decoder consists of four decoder blocks followed by a convolution with a 1x1 kernel and a tanh activation. Each decoder block follows the same general architecture as the decoder blocks of the proposed segmentation model. The number of channels (denoted $C_i$ in figure 4(d)) for the decoder blocks $D_1, D_2, D_3$, and $D_4$ are 256, 128, 64, and 32, respectively.

### Loss Function

To help mitigate issues related to class imbalance, the segmentation model is trained using the Dice[49] loss. Given the ground truth and predicted segmentation masks—$g$ and $p$, respectively—the Dice loss is defined by:

$$L_{\text{Dice}} = 1 - \left(\frac{1}{C}\sum_c^C \frac{2\sum_i^N p_{c,i} \cdot g_{c,i}}{\sum_i^N p_{c,i}^2 + \sum_i^N g_{c,i}^2}\right), \quad (5)$$

where $N$ is the total number of pixels in the mask and $p_{c,i}$ and $g_{c,i}$ are the prediction and ground truth, respectively, for the $i$th pixel in the $c$th class.

To help mitigate the severe imbalance between correctly and incorrectly predicted pixels, the meta-model is trained using a weighted mean squared error (MSE) variant of the Focal-R loss function[50] which we refer to as the weighted focal MSE (WFMSE). The SoftLabel[39] is used to generate the ground truth uncertainty values. Given the ground truth and the predicted uncertainties, denoted $u$ and $\hat{u}$, respectively, the WFMSE loss is:

$$L_{\text{WFMSE}} = \frac{1}{N}\sum_{i=1}^N e_i \cdot \text{SE}(u_i, \hat{u}_i) \cdot \left(2 \cdot \sigma(\beta \cdot \text{AE}(u_i, \hat{u}_i)) - 1\right)^\gamma \quad (8)$$

where $e_i$ is the error weight for the $i$th pixel, SE and AE are the squared and absolute error functions, defined as

$$\text{SE}(u_i, \hat{u}_i) = (u_i - \hat{u}_i)^2 \quad (9)$$

$$\text{AE}(u_i, \hat{u}_i) = |u_i - \hat{u}_i| \quad (10)$$

σ is the sigmoid function, and β, γ are hyperparameters used to control the importance of the absolute error and down weighting of samples with small residuals, respectively.

### Training Strategy

For segmentation model training, we introduced a three-stage training strategy that utilizes natural images and images from related tasks. This sequential training design aids in convergence of the model during the optimization process and mitigates issues related to insufficient training images for this task. At stage one, a ResNet-152 classification model is pretrained on the ImageNet[51] dataset. The pretraining stage enables the model to extract representative image features (e.g., edges, textures, and shapes) for common objects. At stage two, the weights of the backbone from stage one are adopted to initialize the residual module of the proposed segmentation model, then the whole segmentation model is trained using the AGR-2 dataset, a dataset which features images similar to those from this task. This stage tunes the original model to understand the basic image features of TRISO particles and to perform segmentation. All 2,171 particles were used for training. At stage three, the segmentation model is finetuned using images of AGR-5/6/7 particles, which enables the model to adapt to images from this task. The three-stage design defines a coarse-to-fine strategy for gradually improving the performance of the segmentation model, which is more suitable for tasks that lack sufficient training data.

The segmentation model was trained for 50 epochs using an Adam optimizer[52], with a learning rate of 1e-3 and a batch size of 4. The learning rate was reduced by a factor of 0.1 whenever the training loss failed to improve during the most recent five epochs. The images were first converted to grayscale and then normalized. Next, the images and masks were resized to 512×512 by using bilinear and nearest neighbor interpolation, respectively. To mitigate overfitting and encourage generalization to unseen images, extensive augmentation was used during training. Among the augmentation transforms were horizontal and vertical flipping, random changes in brightness and contrast, random shadows, additive noise, grid distortion, elastic transform, and random scaling between 0.8 and 1.2. The flipping, random changes in brightness and contrast, grid distortion, elastic transform, and random scaling were all performed with 50% probability. In addition, contrast-limited adaptive histogram equalization was applied after the random brightness/contrast changes and random shadow operations, to prevent large variance in the image pixels.

Once the segmentation model has completed training, the meta model is trained using the same training set as the segmentation model for 50 epochs using an Adam optimizer with a learning rate of 1e-4 and a batch size of 16. Throughout meta-model training, the segmentation model is frozen. No data augmentation is used. For $L_{\text{WFMSE}}$, $e_i$ is set to 1 and 8 for correctly and incorrectly classified pixels, respectively, and $\beta, \gamma$ are set to 20, 1, respectively. The best model is selected according to the highest Average Precision-Error (described below in the "Uncertainty Quantification Evaluation Setting" subsection) on the validation dataset.

**Experimental Results**

**Data and Segmentation Model Evaluation Setting**

| Training set | Validation set | Test set | Total particles |
|---|---|---|---|
| 326 | 82 | 102 | 510 |

**Table 1.** Sample distribution of the AGR-5/6/7 dataset used for the experiments.

In this work, 510 images of TRISO particles from various compacts were annotated by experienced experts, using the labeling tool described in the Layer Annotation subsection. A total of 326 images were used for training, 82 for validation (e.g., hyperparameter tuning), and 102 for testing. The overall distribution of samples is shown in Table 1. To evaluate model performance, we used the Intersection over Union (IoU), mean IoU (mIoU), precision (P) and mean precision (mP) metrics. The IoU for class $c$ is computed as:

$$\text{IoU}_c = \frac{\text{TP}_c}{\text{TP}_c + \text{FP}_c + \text{FN}_c}, \qquad (11)$$

where $\text{TP}_c$, $\text{FP}_c$, and $\text{FN}_c$ are the true positives, false positives, and false negatives for class $c$, respectively. True positives indicate pixels with the same ground truth and predicted class; false positives indicate pixels labeled as another class which are predicted as class $c$, and false negatives indicate pixels that are labeled as class $c$ but are predicted to belong to another class. The mIoU is defined per:

$$\text{mIoU} = \frac{1}{C} \sum_c^C \text{IoU}_c, \qquad (12)$$

where $C$ is the total number of classes. The precision for class $c$ is:

$$P_c = \frac{\text{TP}_c}{\text{TP}_c + \text{FP}_c}. \qquad (13)$$

Meanwhile, the mean precision is:

$$\mathrm{mP} = \frac{1}{C} \sum_c \mathrm{P}_c. \qquad (14)$$

**Meta-Model Evaluation Setting**

In this work, we evaluate how well the proposed uncertainty model can detect misclassified pixels using two threshold agnostic measures, one error-based measure, and three threshold specific measures. The threshold agnostic measures were the average precision (AP), and AP-error (AP-E). The error-based measure used was the mean squared error (MSE). The threshold specific measures used were specificity (Spec), sensitivity (Sens) and an F1-type metric that is the harmonic mean of specificity and sensitivity[39] (F1-SS).

AP provides an overall summary of the precision recall curve and can be computed as

$$\mathrm{AP} = \sum_n (R_n - R_{n-1}) P_n, \qquad (15)$$

where $R_n$ and $P_n$ are the recall and precision at the $n$th threshold in the precision recall curve and a correct classification represents the positive class. AP-E is computed the same way but using incorrect classifications (i.e. predicted by the segmentation model as the wrong class) as the positive class.

The MSE quantifies how close the model predictions are to the ground truth values and is computed according to:

$$\mathrm{MSE} = \frac{1}{n} \sum_{i=1}^{n} (u_i - \hat{u}_i)^2, \qquad (16)$$

Where $u_i, \hat{u}_i$ are the ground truth and predicted SoftLabel for the $i$-th pixel.

Spec quantifies the percentage of true negatives (correct classifications) detected by the meta-model and is computed according to:

$$\mathrm{Spec} = \frac{TN}{TN + FP}, \qquad (17)$$

where TN and FP represent true negatives and false positives. In this case, true negatives represent pixels labeled as correctly classified that are predicted as correctly classified and false positives represent pixels labeled as incorrectly classified that are predicted as correctly classified. The Sens quantifies the percentage of true positives (incorrect classifications) detected by the meta-model and is computed according to:

$$\mathrm{Sens} = \frac{TP}{TP + FN}, \qquad (18)$$

where TP and FN represent true positive and true negatives. In this case, true positives represent pixels labeled as incorrectly classified that are predicted as incorrectly classified and false negatives represent pixels labeled as incorrectly classified that are predicted as correctly classified. F1-SS provides an overall summary of the Spec and Sens performance and is computed according to:

$$\mathrm{F1-SS} = 2 \cdot \frac{\mathrm{Sens} \cdot \mathrm{Spec}}{\mathrm{Sens} + \mathrm{Spec}}. \qquad (19)$$

For all threshold specific metrics (Spec, Sens, and F1-SS) we find the optimal threshold according to the maximal F1-SS value over the validation dataset. Specifically, we generate thresholds at a 0.01 interval in the range $[-1,1]$ and select the threshold with the highest F1-SS value.

**Overall Segmentation Performance**

The results of the proposed model, both with and without finetuning on the AGR-5/6/7 dataset, are shown in Table 2. Here, the model *without* finetuning refers to the model obtained from stage two of the training process, whereas the model *with* finetuning refers to the model obtained from stage three of the training process. Finetuning significantly improved overall model performance, resulting in a 23.9% and 7.0% relative increase in mIoU and mP, respectively. In particular, the class-level IoUs for the BG, buffer, IPyC, SiC, and OPyC classes show a relative increase of 15.1%, 14.9%, 19.3%, 16.1%, and 141.1%, respectively. Meanwhile, the class-level precisions for the BG, buffer, IPyC, SiC, and OPyC classes show a relative increase of 13.3%, 12.8%, 8.2%, 6.0%, and 5.1%, respectively.

| | IoU | | | | | | mIoU |
|---|---|---|---|---|---|---|---|
| Model | BG | Kernel | Buffer | IPyC | SiC | OPyC | All |
| w/o FT | 0.844 | 0.959 | 0.805 | 0.792 | 0.831 | 0.392 | 0.771 |
| w/ FT | **0.972** | **0.976** | **0.925** | **0.945** | **0.965** | **0.945** | **0.955** |
| | Precision | | | | | | mP |
| Model | BG | Kernel | Buffer | IPyC | SiC | OPyC | All |
| w/o FT | 0.874 | **0.993** | 0.853 | 0.890 | 0.921 | 0.920 | 0.909 |
| w/ FT | **0.990** | 0.983 | **0.962** | **0.963** | **0.976** | **0.967** | **0.973** |

**Table 2.** Performance for the proposed model, both with and without finetuning (FT) on the AGR-5/6/7 dataset. BG: background; IPyC: inner pyrolytic carbon layer; SiC: silicon carbide layer; OPyC: outer pyrolytic carbon layer; All: all image regions. (The best values are in **bold**.)

Figure 5 shows the qualitative results of the model, both with and without finetuning. Without finetuning, the model struggles to accurately segment the OPyC layer, as it was absent in a large number of the AGR-2 images, and hence the poor IoU performance in Table 2. For rows 2 and 3 of Figure 5, the two particles were not well polished, and thus their images show significant noise. The model without finetuning can still accurately segment the kernel and buffer regions from the two images but performs poorly at segmenting other image regions. The results demonstrate that the model without finetuning struggles to adapt to different levels of image quality, whereas the model with finetuning can effectively address this issue.

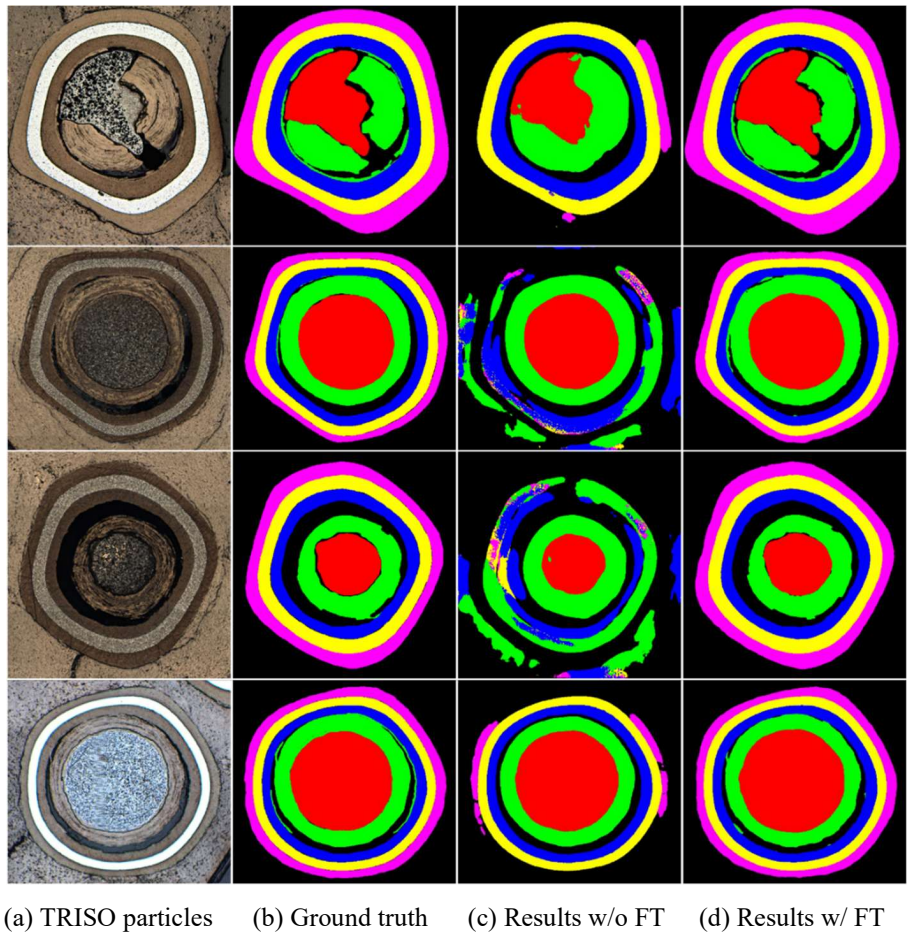

(a) TRISO particles    (b) Ground truth    (c) Results w/o FT    (d) Results w/ FT

**Figure 5.** Comparison of the segmentation results from UA-Net, both with and without finetuning (FT).

### Effectiveness of UA-Net at Segmenting Images of Varying Quality

Figure 6 shows three TRISO images from different compacts. These images vary drastically in terms of quality and appearance, due to different imaging settings and inconsistent material preparation procedures. The first two images show two particles with different color distributions, and the accurate segmentation results in the second and third columns demonstrate that the trained segmentation model was able to adapt to images featuring color variations. The third image shows significant noise, as is usually caused by inappropriate polishing when preparing the materials. This could be highly challenging for conventional image processing approaches as the image texture information has been interrupted, and local image-feature-based approaches could fail. However, the proposed UA-Net can handle the image noise yet still achieve accurate segmentation results.

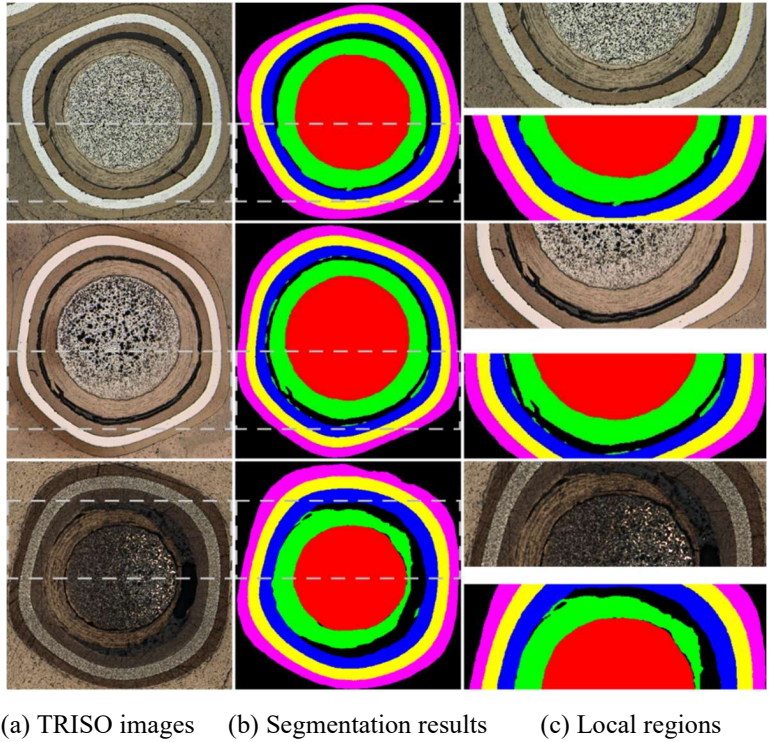

(a) TRISO images     (b) Segmentation results     (c) Local regions

**Figure 6.** Segmentation results obtained by applying the proposed method to varying levels of image quality.

### Misclassification Detection Performance

In this section, we evaluate the misclassification detection performance of the meta-model. We achieve an AP, AP-E, MSE, Spec, Sens, and F1-SS of 0.999, 0.395, 0.128, 0.918, 0.935, and 0.926, respectively. The high Spec and Sens values indicate that the uncertainty model can detect at least 92% of the correctly and incorrectly classified pixels. In critical applications, it is more important to have a high sensitivity as failure to detect incorrect classifications could result in incorrect conclusions being drawn from the results of the segmentation model. The balance between the Spec and Sens can be adjusted by changing the threshold used to determine correct and incorrect classifications. The low AP-E indicates that model performance drops sharply at a low threshold and can most likely be attributed to the extreme imbalance in the dataset used to train the meta-model.

Figure 7 illustrates the ability of the meta-model to identify misclassified regions of the segmentation results produced by the segmentation model of UA-Net. Note that the meta-model predicts soft-labels, which are an indicator of certainty, so we take -$\hat{u}_i$ to obtain the uncertainty maps. In general, there is high uncertainty near borders of the particle layers and the uncertainty correlates well with the error maps shown in column three. Uncertainty near the borders is expected, as these are inherently ambiguous areas of the image. In rows one, two, and four there are small bridges that the segmentation model fails to detect, while row three depicts a polishing artifact that is incorrectly segmented as part of the buffer layer. In all cases, the meta-model produces high uncertainty, signaling a likely misclassification. This indicates that the uncertainty maps could be utilized to alert practitioners to incorrectly segmented areas of the TRISO particle, which could then be corrected before further analysis is performed.

### Comparison Against Other Methods

In this section, we compare UA-Net against RU-Net[7] and six variants of DeepLabV3[15], FCN[11], and MobileNetV3[16]. Specifically, we evaluate the performance of DeepLabV3 models with three feature encoders: MobileNetV3 (MN-DLV3), ResNet-101[46] (RN101-DLV3), and ResNet-50 (RN50-DLV3). We also evaluate FCN-based segmentation with the ResNet-101 (RN101-FCN) and ResNet-50 (RN50-FCN) encoders. The MobileNetV3 model is evaluated using the LR-ASPP segmentation head (MN-LRASPP). All the models were finetuned on the AGR-5/6/7 dataset, using the same training setup as UA-Net.

As shown in Table 3, the proposed UA-Net achieved an mIoU of 0.955 and an mP of 0.973, representing the best average performances when comparing all seven models, and its IoU values in all six classes are the highest. We also observe that RN101-DLV3, RN101-FCN, and RN50-FCN achieve competitive results in many classes, such as the SiC and OPyC regions. The performances indicated by the precision values in Table 3 are mixed. MN-DLV3 obtains the best precision values for the BG and SiC classes, and RN101-DLV3 achieves the best precision values for the IPyC and OPyC classes. The observations imply that the two approaches tend to produce smaller regions than the ground truth. Overall, the differences in the average performance of all the models—apart from MN-DLV3 and MN-LRASPP—are insignificant.

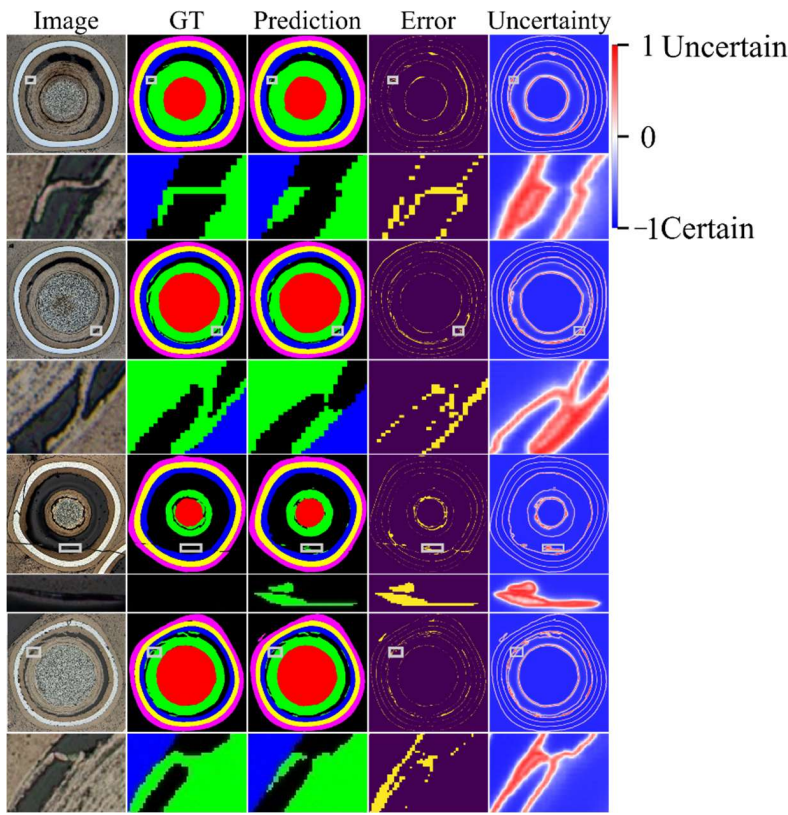

**Figure 7.** Uncertainty results obtained from the meta-model. In the error column, purple and yellow represent pixels that are correctly and incorrectly predicted by the segmentation model, respectively. In the uncertainty column, blue, white, and red represent pixels predicted to have low, medium, and high uncertainty, respectively. The insets below each image show enlarged local details.

| IoU | | | | | | | mIoU |
|---|---|---|---|---|---|---|---|
| Model | BG | Kernel | Buffer | IPyC | SiC | OPyC | All |
| MN-DLV3 | 0.961 | 0.971 | 0.892 | 0.923 | 0.950 | 0.920 | 0.936 |
| RN101-DLV3 | 0.970 | 0.973 | 0.917 | 0.943 | 0.964 | **0.945** | 0.952 |
| RN50-DLV3 | 0.969 | 0.974 | 0.916 | 0.942 | 0.962 | 0.942 | 0.951 |
| RN101-FCN | 0.970 | 0.971 | 0.916 | 0.943 | **0.965** | 0.943 | 0.951 |
| RN50-FCN | 0.970 | 0.974 | 0.917 | 0.944 | **0.965** | 0.944 | 0.952 |
| MN-LRASPP | 0.967 | 0.968 | 0.904 | 0.933 | 0.960 | 0.936 | 0.945 |
| RU-Net | 0.965 | 0.961 | 0.903 | 0.925 | 0.957 | 0.924 | 0.939 |
| UA-Net | **0.972** | **0.976** | **0.925** | **0.945** | **0.965** | **0.945** | **0.955** |
| Precision | | | | | | | mP |
| Model | BG | Kernel | Buffer | IPyC | SiC | OPyC | All |
| MN-DLV3 | **0.990** | 0.983 | 0.929 | 0.945 | **0.989** | 0.929 | 0.961 |
| RN101-DLV3 | 0.989 | 0.981 | 0.955 | **0.964** | 0.974 | **0.969** | 0.972 |
| RN50-DLV3 | 0.989 | **0.985** | 0.956 | 0.963 | 0.974 | 0.964 | 0.972 |
| RN101-FCN | 0.988 | 0.983 | 0.959 | 0.963 | 0.977 | 0.965 | 0.972 |
| RN50-FCN | 0.988 | 0.985 | 0.957 | 0.962 | 0.977 | 0.966 | **0.973** |
| MN-LRASPP | 0.988 | 0.979 | 0.956 | 0.949 | 0.974 | 0.962 | 0.968 |
| RU-Net | 0.985 | 0.967 | **0.962** | 0.951 | 0.967 | 0.953 | 0.972 |
| UA-Net | **0.990** | 0.983 | **0.962** | 0.963 | 0.976 | 0.967 | **0.973** |

**Table 3.** Performance of the proposed UA-Net model as compared to seven other methods also applied to the AGR-5/6/7 dataset. BG: background; IPyC: inner pyrolytic carbon layer; SiC: silicon carbide layer; OPyC: outer pyrolytic carbon layer. (The best values are in **bold**.)

Figure 8 illustrates the capability of UA-Net and the other methods to identify fine details such as narrow gaps and small connections between regions. Accurate segmentation of the fine details in TRISO images is essential for further analysis of the defect types. The image regions below each whole TRISO image show enlarged local details within the grey bounding box. The first two examples display two images, with a small connection between the buffer and IPyC layers. UA-Net correctly identifies both these connections, whereas the other seven methods can, at best, find only one or tend to under segment them. The third example shows an image with a narrow gap between two buffer fragments. RN-DLV3, MN-DLV3, and MN-LRASPP fail to identify the gap, and connect the two buffer fragments together, making it impossible to correctly identify the buffer defects. While RN101-FCN, RN101-DLV3, and RU-Net can separate the two fragments, the gaps generated by RN101-FCN and RN-101-DLV3 are significantly narrower than the original one seen in the ground truth image and RU-Net generates a gap that is larger than the one seen in the ground truth image. The proposed UA-Net not only separates the two fragments, but also accurately recovers the gap.

## Discussion

Developing advanced segmentation approaches for high-resolution image segmentation is a relatively new topic, but notable progress has already been made. However, it is still a challenge to accurately segment fine details in images. For example, in the TRISO images, the narrow bridges and gaps between two layers are difficult to identify, leading to challenges in detecting some types of material defects. The proposed approach achieved promising results by accurately identifying all four layers plus the kernel, improved the fine-detail segmentation, and demonstrated its generalization to images of varying quality.

One drawback of the proposed method is that when trained on images with a specific magnification level, it cannot adapt well to different image scales, as all images are downsampled to a fixed size (in this work, 512×512). This could lead to difficulty accurately segmenting fine details, especially at lower magnification levels (e.g., 10×). One possible solution may be to determine an optimal image resolution/magnification level by experimentation, and then to collect all future images by using that same magnification level. On the other hand, high-magnification images could present challenges due to a difference in spatial context. A simple solution would be to use images from multiple scales in the training dataset. However, this would necessitate labelling a larger amount of data and may therefore be prohibitively expensive. In contrast, a more effective solution would be to adapt techniques from work done in high-resolution remote sensing, where separate model branches are utilized to generate and fuse features with global and local context[53–56]. Another possibility is to use adaptive refinement methods (proposed for high-resolution remote sensing or natural image segmentation) that apply postprocessing modules to coarse segmentation outputs to generate finer, more accurate segmentation results[57–60].

Another issue with the current model is determining the optimal magnification to use for inference. While the model generalizes well to magnification levels of 10× and 20×, higher levels could lead to significant loss of fine detail in the input image, due to extensive downsampling prior to being input to the model. In the future, we plan to investigate how well the model can generalize to additional magnification levels (e.g., 5×, 30×, and 50×), the optimal magnification level for training and inference, and the possible trade-offs that exist.

While this work applies the proposed segmentation model to TRISO-based fuels, the model could easily be adapted to other fuel types, materials, or image modalities. This would involve collecting and annotating an appropriate dataset and changing the output layer according to the number of image features of interest. For example, if the material only contains two classes of interest, the output layer would need to be updated to predict three classes (two for the materials of interest and one for background). If the new fuel type or material is similar to AGR-5/6/7 in terms of geometry and image texture, UA-Net could be fine-tuned using a small dataset. Otherwise, a slightly larger dataset could be necessary to obtain good results.

Here, we apply our framework to TRISO particle segmentation. However, it is a general framework that can easily be adapted and applied to quantify the uncertainty of segmentation models with little overhead thereby allowing practitioners the ability to detect erroneous predictions and intervene in the segmentation process. This is particularly important for deployment in critical applications where incorrect model predictions can have severe consequences. Moreover, due to the post-hoc nature of the uncertainty component, our framework can be applied in areas where an already trained segmentation model is available. In the future, we plan to adapt this framework for other segmentation problems, those in critical applications.

## Conclusion

In this work, we propose UA-Net, a novel uncertainty-aware framework for segmenting the images of TRISO particles. We also propose a novel three-stage training process for training the segmentation model. The training process significantly enhances segmentation performance across different TRISO particle image regions and requires substantially fewer computational resources than large-scale pretraining on natural images. The proposed segmentation model can accurately segment particles of varying image quality and accurately capture fine details, thereby outperforming other methods. On the other hand, the meta-model can detect most misclassified regions of the segmentation prediction, thereby allowing human intervention in the segmentation process. In the future, we plan to investigate methods for high-resolution image segmentation to further improve the segmentation of fine details as well as applying our uncertainty framework to other segmentation problems.

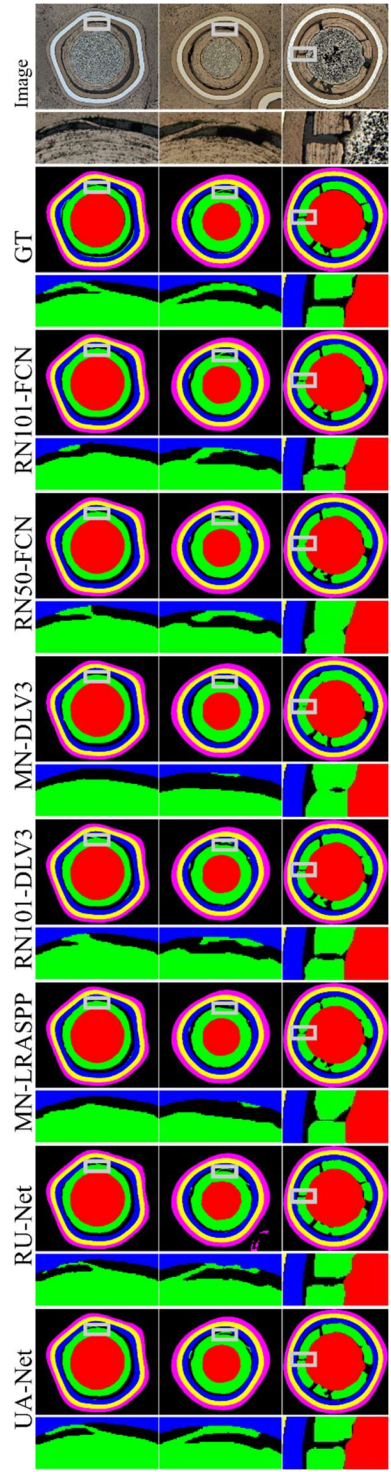

**Figure 8.** Segmentation results obtained from the proposed UA-Net model and the seven other models. The insets below each image show enlarged local details.

### Acknowledgments

The authors gratefully acknowledge the financial support from the U.S. Department of Energy, Advanced Fuels Campaign (AFC) of the Nuclear Technology Research and Development program in the Office of Nuclear Energy. This manuscript has been authorized by Battelle Energy Alliance, LLC, under Contract No.DE-AC07–05ID14517 with the U.S. Department of Energy. The authors also extend their sincere gratitude to the staff at the Hot Fuel Examination Facility (HFEF) of Idaho National Laboratory for their invaluable assistance with sample preparation, handling, and data collection.